\documentclass[runningheads]{llncs}

\usepackage[T1]{fontenc}
\usepackage{svg}
\usepackage{graphicx}
\usepackage{pdfpages}
\usepackage{subcaption}
\usepackage{comment}
\usepackage{amsmath}
\usepackage{booktabs}
\usepackage{caption}
\usepackage{wrapfig}
\usepackage{algorithmic}
\usepackage{algorithm}
\usepackage{grffile}
\usepackage{hyperref}

\usepackage{float}
\begin{document}
\title{Advancing Brain Tumor Segmentation via Attention-based 3D U-Net Architecture and Digital Image Processing}

\author{Eyad Gad\inst{1}\and Seif Soliman\inst{1}\and\ M. Saeed Darweesh\inst{1,2}}

\authorrunning{E. Gad et al.}
\titlerunning{Advancing Brain Tumor Segmentation}
\institute{School of Engineering and Applied Sciences, Nile University, Giza, 12677, Egypt\and Wireless Intelligent Networks Center (WINC), Nile University, Giza, 12677, Egypt\\
\email{\{\href{mailto:e.gad@nu.edu.eg}{e.gad},
\href{mailto:s.soliman@nu.edu.eg}{s.soliman},
\href{mailto:mdarweesh@nu.edu.eg}{mdarweesh}\}@nu.edu.eg}}
\maketitle    

\begin{abstract}
In the realm of medical diagnostics, rapid advancements in Artificial Intelligence (AI) have significantly yielded remarkable improvements in brain tumor segmentation. Encoder-Decoder architectures, such as U-Net, have played a transformative role by effectively extracting meaningful representations in 3D brain tumor segmentation from Magnetic resonance imaging (MRI) scans. However, standard U-Net models encounter challenges in accurately delineating tumor regions, especially when dealing with irregular shapes and ambiguous boundaries. Additionally, training robust segmentation models on high-resolution MRI data, such as the BraTS datasets, necessitates high computational resources and often faces challenges associated with class imbalance. This study proposes the integration of the attention mechanism into the 3D U-Net model, enabling the model to capture intricate details and prioritize informative regions during the segmentation process. Additionally, a tumor detection algorithm based on digital image processing techniques is utilized to address the issue of imbalanced training data and mitigate bias. This study aims to enhance the performance of brain tumor segmentation, ultimately improving the reliability of diagnosis. The proposed model is thoroughly evaluated and assessed on the BraTS 2020 dataset using various performance metrics to accomplish this goal. The obtained results indicate that the model outperformed related studies, exhibiting dice of 0.975, specificity of 0.988, and sensitivity of 0.995, indicating the efficacy of the proposed model in improving brain tumor segmentation, offering valuable insights for reliable diagnosis in clinical settings.


\end{abstract}

\section{Introduction}

A brain tumor is a collection of abnormal cells that grows in the brain or central spine canal. These cells undergo a process called mitosis, where a single cell duplicates its entire contents, including chromosomes, and divides into two identical daughter cells. Consequently, an abnormal mass, which can be either non-cancerous (benign) or cancerous (malignant), forms. Depending on their location and size, brain tumors can manifest a range of symptoms, such as headaches, seizures, cognitive decline, mood alterations, and impaired movement or coordination. Brain tumors can affect people of all ages, but they are more common in older adults. According to the Global Cancer Statistics, there are approximately 300,000 new cases of brain tumors diagnosed globally each year \cite{Intro1}. Thus, early detection and treatment play a crucial role in managing brain tumors and enhancing patients' quality of life.

Recent advancements in AI have revolutionized the field of medical imaging, particularly in the domain of brain malignancies. AI technologies have significantly improved the segmentation, identification, and survival prediction of tumors and other diseases \cite{Intro2,Intro3,Intro4}. By accurately delineating brain tumors from medical images, AI facilitates early diagnosis, leading to better prognostic capabilities and informed decision-making by medical professionals. Furthermore, AI algorithms enable the tracking of tumor changes over time, aiding in disease monitoring and providing real-time support during surgical interventions \cite{Intro4,Intro5}. The automation of image analysis through AI not only reduces the risk of errors but also enhances efficiency within the healthcare system. Additionally, AI has the potential to facilitate remote consultations, reducing the need for in-person visits and lowering associated costs. As AI continues to advance, its transformative impact on brain tumor segmentation and patient care will continue to expand.

In the past, classical machine learning techniques were commonly employed for segmentation tasks until the advent of deep learning techniques, such as CNNs, revolutionized the field. CNNs, specifically, have been widely adopted for MRI segmentation, with a focus on brain tumor segmentation. Architectures like U-Net have demonstrated high performance by effectively capturing local features and extracting meaningful representations from MRI scans \cite{Intro6}. While traditional U-Net models have achieved considerable success in numerous studies, they may encounter challenges when accurately distinguishing between tumor regions and healthy brain tissue, especially in cases where tumors have irregular shapes or indistinct boundaries. This can lead to imprecise segmentation outcomes, making it difficult to precisely locate tumors. Furthermore, the computational requirements for brain tumor segmentation are noteworthy, as achieving accurate segmentation heavily relies on substantial computational power for training U-Net models on high-resolution MRI scans. Scaling up these models to handle large datasets and effectively testing novel concepts and adjustments exacerbates these challenges \cite{Intro7,Intro10,Intro11}. 

Another challenge arises from the imbalanced distribution of pixels across the different classes, with some classes having significantly fewer samples compared to others. This class imbalance complicates the training process as the model tends to prioritize the majority class, leading to biased results. The limited representation of minority classes can result in their underestimation or misclassification during the segmentation process.

The study proposes the integration of the attention mechanism into the U-Net model to mitigate the challenges in brain tumor segmentation. By incorporating the attention mechanism, the model can selectively focus on relevant regions and prioritize important features during segmentation \cite{Intro12}, resulting in improved differentiation between tumor regions and healthy brain tissue, even in scenarios involving irregular tumor shapes or unclear boundaries. To address the class imbalance, a tumor detection method based on digital image processing is employed as a data preparation step to detect tumors in scans, which balances the classes prior to training and reduces bias in the segmentation results.

To sum up, this study makes two key contributions. Firstly, it integrates the attention mechanism into the U-Net model. Secondly, it employs a tumor detection method based on digital image processing techniques to address class imbalance and reduce bias in the segmentation results. These contributions aim to enhance the accuracy and reliability of brain tumor segmentation, leading to improved diagnosis and treatment planning.

\section{Literature Review}

Over the past few years, several approaches have been proposed for brain tumor segmentation, ranging from traditional image-processing techniques to deep learning-based methods. Throughout the upcoming literature review, an overview is provided containing recent advances in brain tumor segmentation methods and techniques, highlighting their strengths and limitations. 

Starting with Montaha et al. \cite{Lit1}, which proposed a 2D U-Net implementation for brain tumor segmentation that uses a single slice of 3D MRI to minimize computational cost while achieving high performance. MRI intensity normalization is employed as a pre-processing technique, converting minimum values to 0 and maximum values to 1. Rescaling is applied to the middle single slices instead of all 155 slices, reducing computational complexity. The segmentation pipeline uses a 2D U-Net architecture with a compact encoder for feature extraction and a decoder for image reconstruction, incorporating skip connections to reduce information loss and address the vanishing gradient problem. The model is separately trained and validated on four MRI modalities and labeled segmented ROI images. The proposed model achieved a dice score of 0.9386, accuracy of 0.9941, and sensitivity of 0.9897 on the BraTS2020 dataset, demonstrating its potential for clinical application.

Another recently proposed approach from Ilhan et al. \cite{Lit2} introduces a brain tumor segmentation system that uses non-parametric tumor localization and enhancement methods with a U-Net implementation. The study heavily relies on pre-processing techniques and a U-Net architecture, and the data preparation modules used for this study comprise 2D axial images obtained from patients' FLAIR modalities. To determine the background and tumorous regions in brain MRI scans, image histograms are obtained, and a non-parametric threshold value is calculated using the frequency of intensity values. In low-contrast MRI scans, the standard deviation is used to enhance the distinguishability and contrast between the background and tumorous regions, allowing easy localization of tumors. The enhanced images are fed into the U-Net architecture for segmentation, where the contraction path (encoder) and expansion path (decoder) enable the architecture to learn fine-grained details. The system achieved outstanding segmentation performance on three benchmark datasets, with dice scores of 0.94, 0.87, and 0.88 for BraTS2012 HGG-LGG, BraTS2019, and BraTS2020, respectively.

Moreover, N. Cinar et al. \cite{Lit3} developed a novel hybrid architecture with a pre-trained DenseNet121 and U-Net architecture to identify multiple tumor sub-regions for the segmentation process. In terms of data preparation, MRI images were enhanced using clipping and cropping into a (64$\times$64) size whilst centering the tumor in the image. The images were also divided into smaller parts to shorten the training period. The Otsu threshold technique was also applied as it’s tasked to differentiate between-class variance of foreground (tumor) and background pixels, and z-score normalization was used to ensure homogeneity. The architecture of the proposed model combines the DenseNet121 architecture, which is pre-trained on ImageNet, and a U-Net architecture implementation. DenseNet121 network was used as the encoder which is responsible for feature extraction, and the fully connected layer was removed and replaced with a decoder. Skip connections were used to transmit the attributes of the input image to the decoder layers. The segmented tumor areas are concatenated together in order to obtain the fully segmented image. This model was validated on the BraTS2019 dataset whilst disregarding all T1 scans due to lack of resolution. Dice coefficient obtained for the sub-regions are as follows WT 0.959, CT 0.943 and ET 0.892.

Lastly, Raza et al. \cite{Lit4} proposed architecture, referred to as the Deep Residual U-Net (dResU-Net), which is an end-to-end encoder-decoder-based model that utilizes residual blocks with shortcut connections in the encoder part of the U-Net model. The authors utilized image standardization and normalization on all MRI images, resizing them to 128$\times$128$\times$128 dimensions and stacking them together to form a 128$\times$128$\times$128$\times$4 input. The proposed architecture utilized a U-Net with residual blocks in the encoder part to extract low and high-level features, overcoming the vanishing gradient problem whilst transferring feature maps of each encoder level to its corresponding decoder level using the skip connection. The bottleneck and decoder parts contain plain convolutional blocks for predicting segmentation masks, while the expanding path recovers the image into its original shape using the traditional up-sampling functions. The proposed model was cross-validated on an external dataset to evaluate its robustness, achieving a Dice Score of 0.8357, 0.8660, and 0.8004 for BraTS 2020 CT, WT, and ET, respectively.  

The recent advancements in brain tumor segmentation techniques, particularly in U-Net variants, have become an increasingly popular choice for brain tumor segmentation due to their ability to effectively learn and extract features from medical images. The various modifications and improvements made to the original U-Net architecture have addressed some of its limitations, such as the vanishing gradient problem, and have resulted in improved segmentation performance. 

\section{Materials and Methods}

\subsection{Proposed Approach}
\begin{figure}
\centering
  \includegraphics[width=0.8\textwidth,height=10pc]{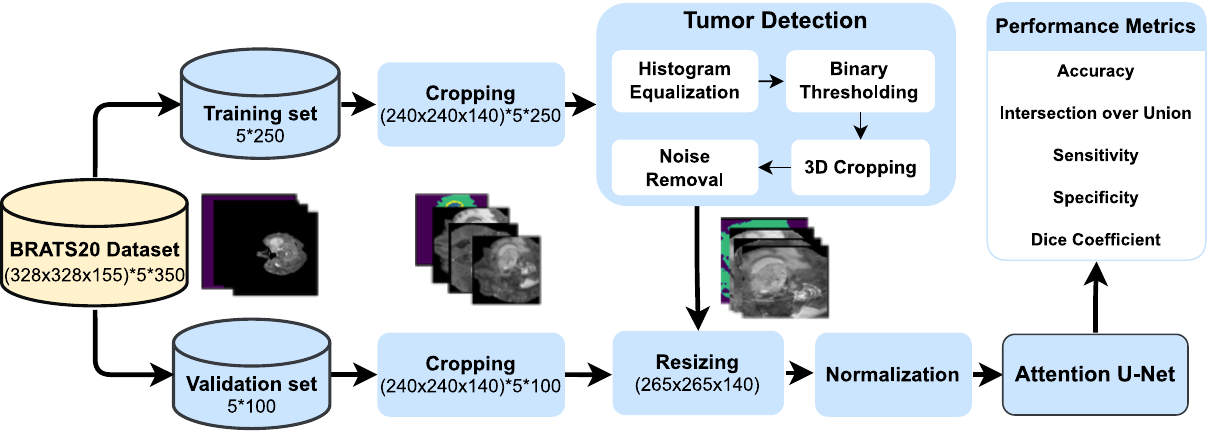}
  \caption{The pipeline of the proposed approach for 3D BraTS}\label{fig:pipeline}
\end{figure}

This paper presents a novel approach for brain tumor segmentation, as illustrated in Figure \ref{fig:pipeline}. The proposed method involves several steps, starting with the partitioning of MRI scans from the BraTS dataset into training and validation sets. Preprocessing techniques, such as cropping, resizing, and normalization, are applied, along with a tumor detection algorithm based on digital image processing, which focuses on the tumor region to improve class balancing within the training set. Subsequently, the attention U-Net model is then trained on the preprocessed data and evaluated using performance metrics. This comprehensive approach aims to achieve precise and efficient brain tumor segmentation.

\subsection{Dataset Sources and Preparation}

In the BraTS dataset, brain tumor segmentation involves classifying pixels into distinct classes representing three non-overlapping subregions: edema, enhancing tumor, and necrotic core or non-enhancing tumor (NCR/NET). These subregions have different biological properties and are crucial for tumor characterization. In this study, three commonly used regions of interest (ROIs) or classes based on these subregions are utilized.

The first ROI of interest is the whole tumor (WT), which includes all three subregions and provides a comprehensive representation of the tumor's extent. The second ROI is the tumor core (TC), focusing on the NCR/NET and enhancing tumor subregions for characterization and treatment planning. Lastly, the third ROI is the enhancing tumor (ET), targeting the area that shows enhancement and providing important information for clinical decisions.

In this study, the BraTS2020 dataset was examined, which consists of 350 MRI scans specifically focusing on glioma brain tumors. These scans encompass four different modalities, namely T1-weighted (T1), T1-weighted with contrast enhancement (T1ce), T2-weighted (T2), and Fluid-Attenuated Inversion Recovery (FLAIR). Brain tumors can emerge in various regions of the brain and can vary significantly in terms of their size and shape. Additionally, the intensity of tumor tissue can overlap with that of healthy brain tissue, presenting a significant challenge in distinguishing between the two. For instance, in T1 MRI scans, a bright tumor border might be visible, while the tumor area itself might be highlighted in T2 MRI scans. Moreover, FLAIR MRI scans assist in differentiating edema from cerebrospinal fluid. To tackle this challenge, a fundamental approach involves integrating data from multiple MRI modalities, such as T1, T1ce, T2, and FLAIR. By combining information from these different modalities, it becomes possible to obtain a more comprehensive and accurate representation of the tumor and its surrounding tissues. This multi-modal approach improves the ability to differentiate between tumor regions and healthy brain tissue, enhancing the overall accuracy of tumor segmentation. 

\begin{algorithm}
\caption{Tumor Detection Algorithm}\label{alg:tumor_detection}
\begin{algorithmic}[1]
  \STATE \textbf{Input}: 3D MRI Scan $V$ of size $m \times n \times s$
  \STATE \textbf{Output}: Coordinates of Detected Tumor $largest\_coor$
  \STATE $coors \gets$ Empty list
  \FOR{$i=1$ to $s$}
      \STATE \textbf{Thresholding:} $k\gets$ threshold($V[i]$,$thresh$)
      
      \STATE \textbf{Noise Removal:} Dilate($k$) to connect nearby objects
      \FOR{each $object$ in $k$}
          \IF{area of $object < \text{area\_thresh}$}
              \STATE Eliminate $object$
          \ENDIF
      \ENDFOR
      \STATE $coors \gets$ coordinates of square points containing large objects in $k$
  \ENDFOR
  \STATE \textbf{Tumor Cropping:}
  \STATE $largest\_area \gets$ area of $coors[0]$
  \STATE $largest\_coor \gets coors[0]$
  \FOR{$cur\_coor=1$ to len($coors$)}
     \IF{area of $coors[cur\_coor] > largest\_area$ \textbf{and} region of $coors[cur\_coor]$ includes region of $largest\_coor$}
                \STATE $largest\_area \gets$ area of $coors[cur\_coor]$
                \STATE $largest\_coor \gets coors[cur\_coor]$
      \ENDIF
  \ENDFOR
\end{algorithmic}
\end{algorithm}

\begin{figure}
  \centering
  \begin{subfigure}{0.45\textwidth}
    \includegraphics[width=\linewidth]{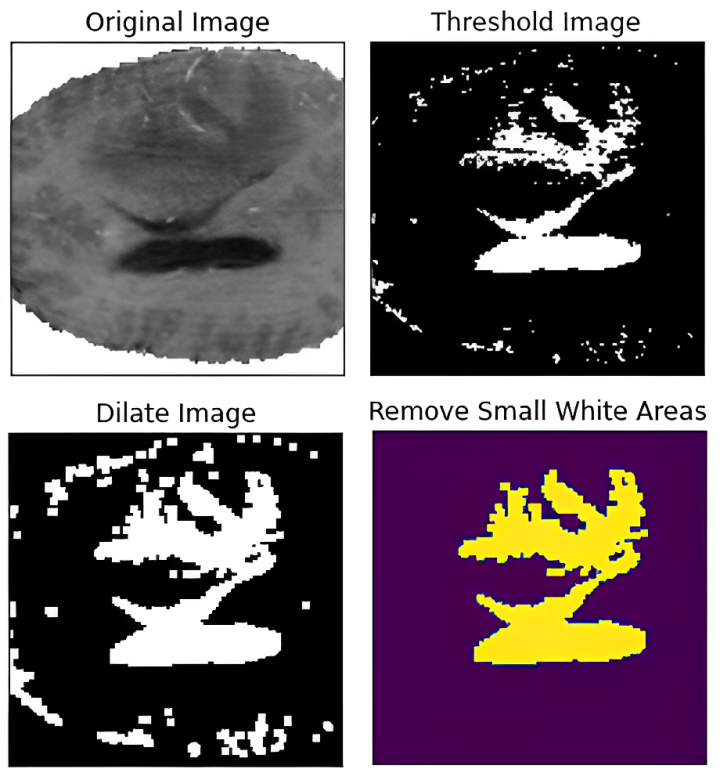}
    \caption{}
    \label{fig:accuracy}
  \end{subfigure}
  \begin{subfigure}{0.45\textwidth}
    \includegraphics[width=\linewidth]{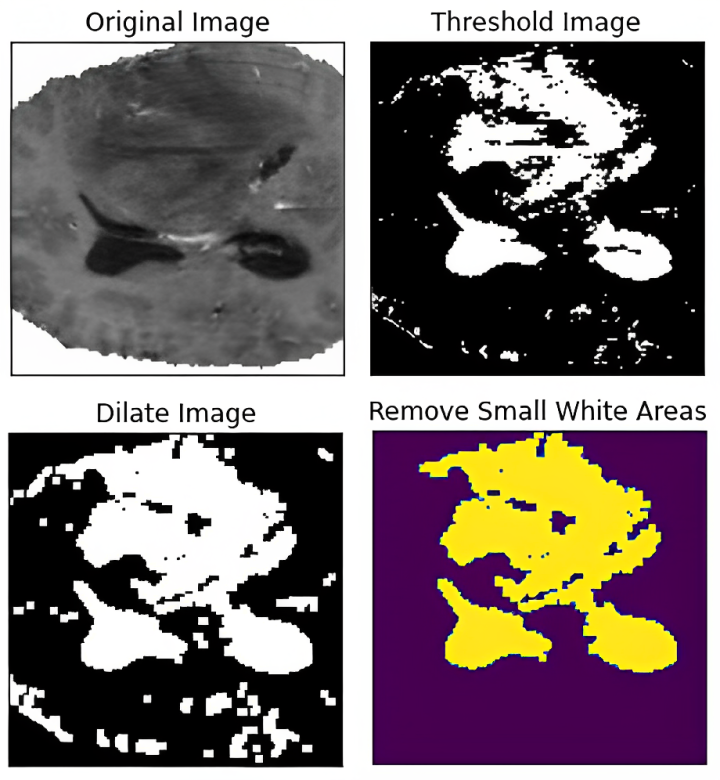}
    \caption{}
    \label{fig:dice_loss}
  \end{subfigure}
  \caption{Tumor detection results}
  \label{fig:tumordetection}
\end{figure}

Training on a dataset of 250 scans, each consisting of 5 dimensions (4 modalities and a mask), with each dimension having 328$\times$328$\times$155 pixels, would be inefficient and impractical. To address this, preprocessing procedures are undertaken to eliminate irrelevant areas such as the background and unaffected healthy tissues, and enhance tumor detection. The first step involves cropping the scans and their corresponding masks to constrain them to the brain region defined by assigned coordinates, ensuring that subsequent analysis focuses only on the relevant areas for tumor detection. Next, a tumor detection algorithm is applied, utilizing multiple image processing techniques to enhance accuracy. The algorithm begins with histogram equalization, which improves the contrast and visibility of tumor regions by making darker and brighter regions more distinguishable, making them more prominent and easier to detect. The algorithm, as illustrated in Algorithm \ref{alg:tumor_detection}, employs thresholding to differentiate between tumor and non-tumor regions based on a specific intensity threshold. Pixels with intensities above the threshold are classified as part of the tumor region and vice versa. However, it is important to note that fully distinguishing the tumor region can be challenging due to healthy regions with tumor-like intensities, leading to false positives. As shown in Figure \ref{fig:tumordetection}, small objects considered noise in comparison to the large tumor object.

To address noise and refine tumor detection further, noise removal techniques are applied. Initially, dilation is employed to connect nearby large objects to the tumor region and create thickness in the boundary. This ensures that when the detected tumor is cropped, the entire tumor is included within the cropped object. Subsequently, the algorithm iterates through the objects and eliminates those with small areas, effectively removing small false positives or noise that may have been detected.

After applying these steps to each 2D scan in the MRI volume and obtaining the detected tumor for each scan, the next stage involves cropping the tumor to encompass the entire volume. This is achieved by determining the coordinates that encapsulate the tumor throughout the entire MRI volume. The algorithm iterates through the detected tumor coordinates obtained from each 2D scan and identifies the largest area. By examining the coordinates of this largest area, it can be determined whether they encompass all the other detected tumor coordinates. This ensures that the resulting cropping includes the entire tumor within the MRI volume, providing a comprehensive representation of the tumor across multiple scans. Lastly, all the cropped scans are resized and normalized, preparing them for training.

\subsection{Attention U-Net}
\begin{figure}
\centering
\includegraphics[width=1\textwidth,height=12pc]{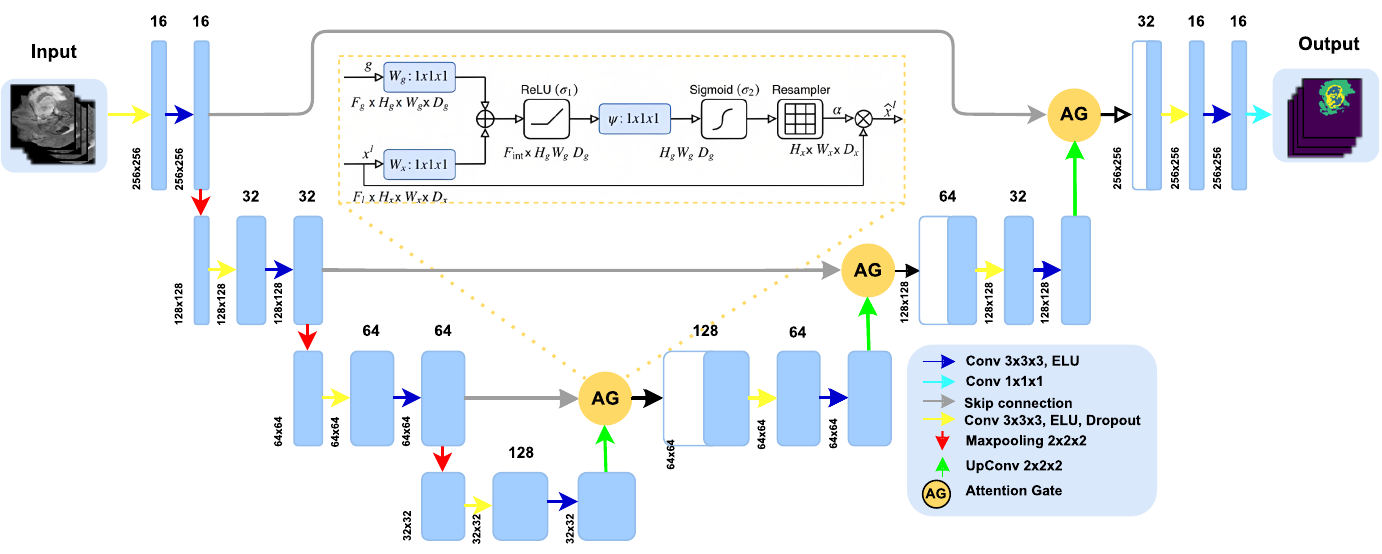} 
\caption{The architecture of 3D U-Net model with attention gate}\label{fig:atten}
\end{figure}

In this study, the attention U-Net architecture was utilized for segmenting brain tumor. The attention U-Net consists of a contracting path, which includes convolutional and max pooling layers, followed by an expanding path with convolutional and up-sampling layers. What sets the attention U-Net apart from a standard U-Net is the incorporation of an attention mechanism or attention gate in the skip connections, as it improves the network’s ability to focus on relevant features while disregarding irrelevant ones, thereby enhancing the accuracy of tumor segmentation.
As illustrated in Figure \ref{fig:atten}, the attention gate involves two input vectors: X and G. Vector G is derived from the lower layer of the network and has smaller dimensions but better feature representation. Vector X undergoes a strided convolution, while vector G undergoes a 1$\times$1 convolution. These vectors are then combined through element-wise summation, resulting in a new vector. This new vector is passed through activation and convolutional layers to reduce its dimensions. A sigmoid layer is then applied to produce attention coefficients, which represent the importance of each element in vector X. To restore the attention coefficients to the original dimensions, trilinear interpolation is used. The attention coefficients are multiplied with the original vector X, scaling it based on relevance. Finally, the modified vector is passed through the skip connection for further processing.

The brain tumor training procedure of the U-Net model included the utilization of the dice loss function given by:
\begin{equation}
\text{Dice Loss} = 1 - \frac{2 \cdot \sum_{i=1}^{N} p_i \cdot t_i}{\sum_{i=1}^{N} p_i^2 + \sum_{i=1}^{N} t_i^2},
\end{equation}
where $p_i$ and $t_i$ are the predicted and target values for each pixel $i$, and $N$ is the total pixels. The Adam optimizer with a learning rate of 0.0001 was used, and a batch size of 16 was employed during training. The model underwent training for 100 epochs, and its performance in tumor segmentation were evaluated.

\subsection{Performance Metrics}

To evaluate the model's performance, various metrics, including Accuracy, Dice Coefficient, Intersection over Union (IoU), Sensitivity, and Specificity were employed. These metrics were assessed during both the training and validation phases using a confusion matrix that includes true positives (TP), false positives (FP), false negatives (FN), and true negatives (TN).

The Dice Coefficient is a widely-used metric to measure pixel-level agreement between predicted segmentation and ground truth. It is computed by subtracting the Dice Loss from 1. Additionally, the IoU metric, also known as the Jaccard index, quantifies the overlap between predicted and true positive regions \cite{Everingham_Van_Gool_Williams_Winn_Zisserman_2009}. It is calculated as:
\begin{equation}
\text{IoU} = \frac{\text{TP}}{\text{TP} + \text{FP} + \text{FN}}
\end{equation}

Additionally, Sensitivity, also known as recall or true positive rate, measures the proportion of actual positives correctly identified by the model. It can be calculated using the following formula:
\begin{equation}
\text{Sensitivity} = \frac{\text{TP}}{\text{TP} + \text{FN}}
\end{equation}
Furthermore, Specificity evaluates the proportion of actual negatives correctly identified by the model and is computed as:
\begin{equation}
\text{Specificity} = \frac{\text{TN}}{\text{TN} + \text{FP}}
\end{equation}
These performance metrics provide valuable insights into the effectiveness and accuracy of the model in accurately capturing true positive regions.

\section{Experimental Results}

 The training process was carried out in two rounds, each consisting of 50 epochs, determined through empirical evaluation and prior research, ensuring sufficient training iterations for convergence and optimal performance. In the first round, a batch size of 16 was used during training, allowing for potentially expediting the convergence process. However, to ensure more precise updates to the model's parameters and potentially improve its performance further, a smaller batch size of 8 was employed in the second round.
\begin{figure}
  \centering
  \begin{subfigure}{0.3\textwidth}
    \includegraphics[width=\linewidth]{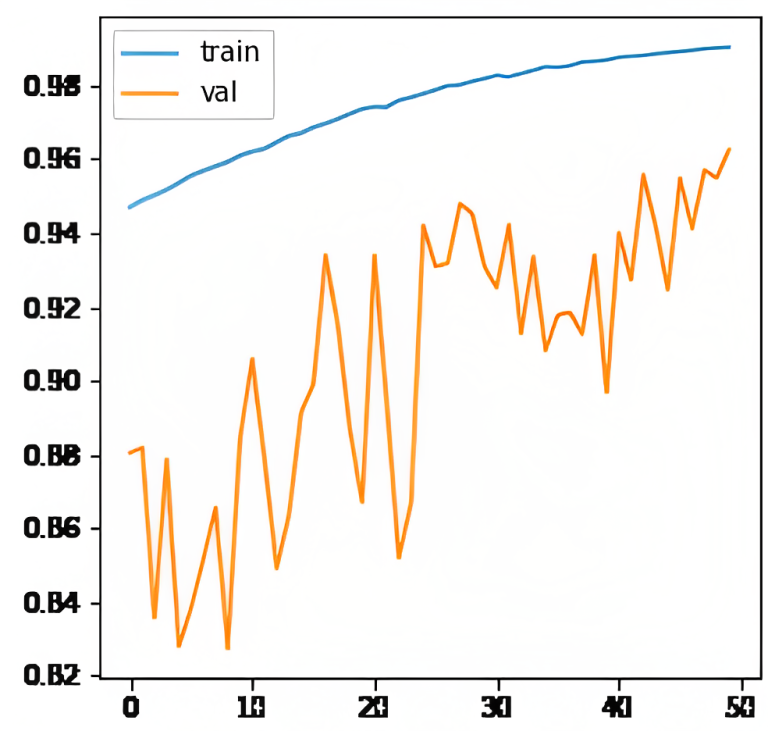}
    \caption{Accuracy}
    \label{fig:accuracy1}
  \end{subfigure}
  \begin{subfigure}{0.3\textwidth}
    \includegraphics[width=\linewidth]{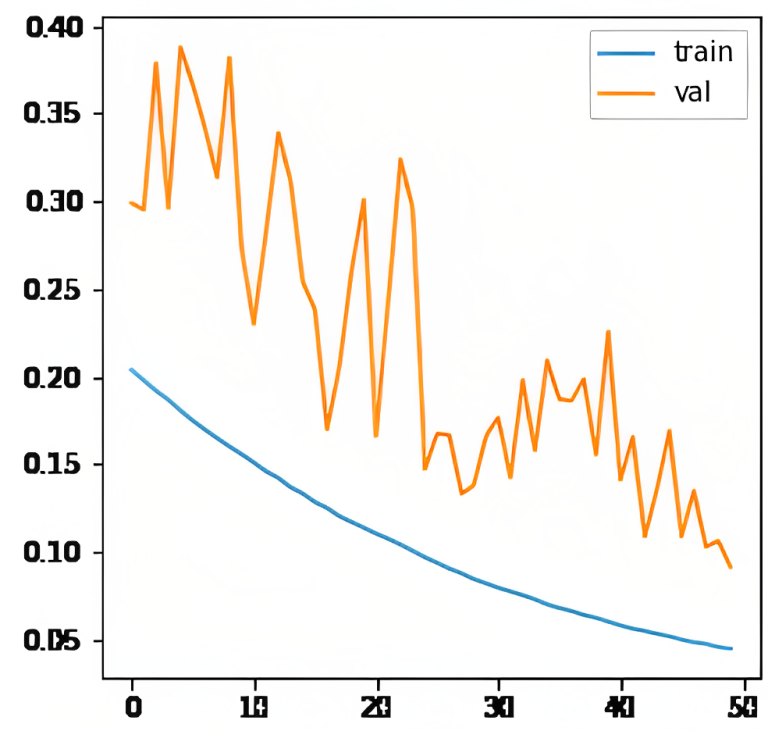}
    \caption{Dice Loss}
    \label{fig:dice_loss1}
  \end{subfigure}
  \begin{subfigure}{0.3\textwidth}
    \includegraphics[width=\linewidth]{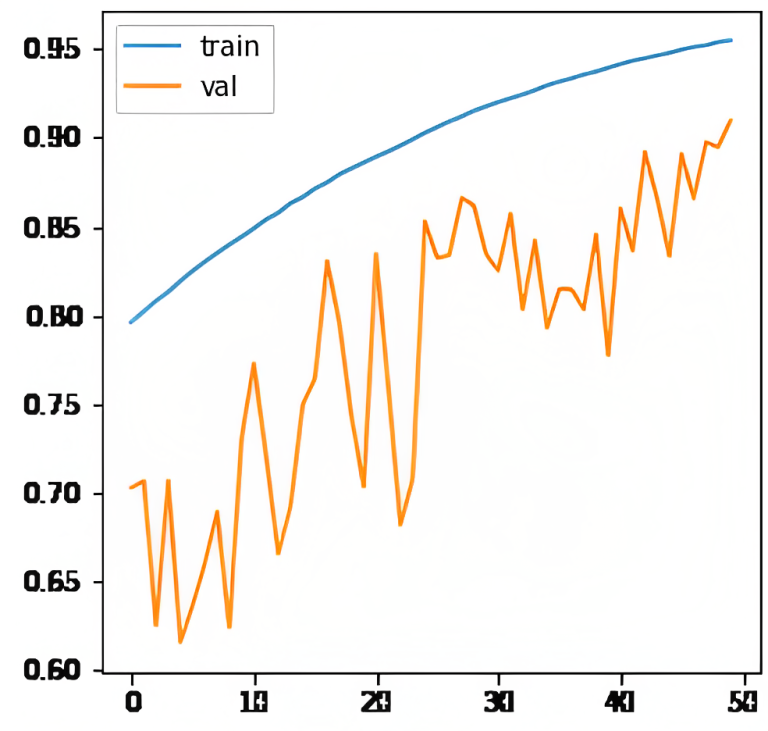}
    \caption{Dice Coefficient}
    \label{fig:dice_coefficient1}
  \end{subfigure}
  \medskip
  \begin{subfigure}{0.3\textwidth}
    \includegraphics[width=\linewidth]{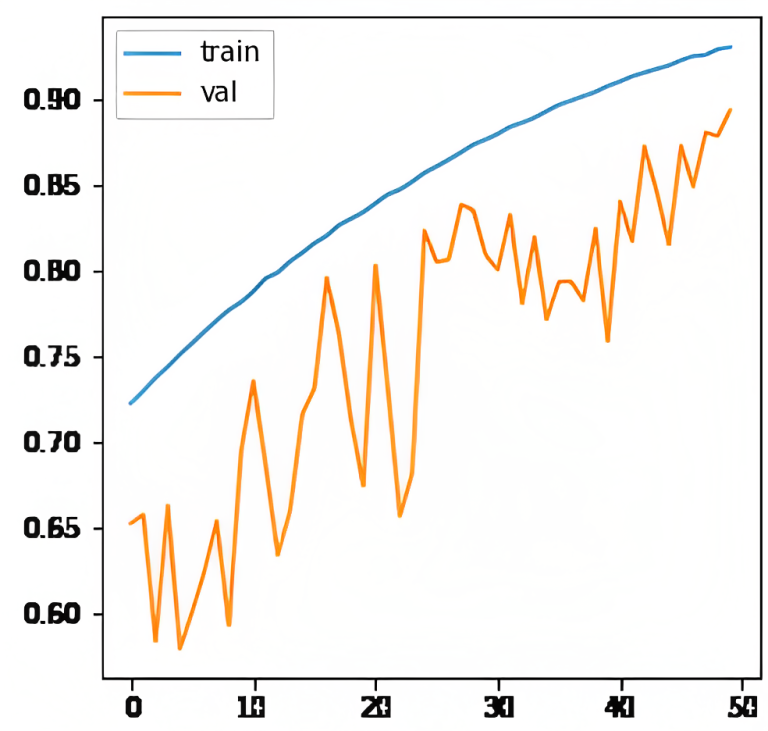}
    \caption{IoU}
    \label{fig:iou1}
  \end{subfigure}
  \begin{subfigure}{0.3\textwidth}
    \includegraphics[width=\linewidth]{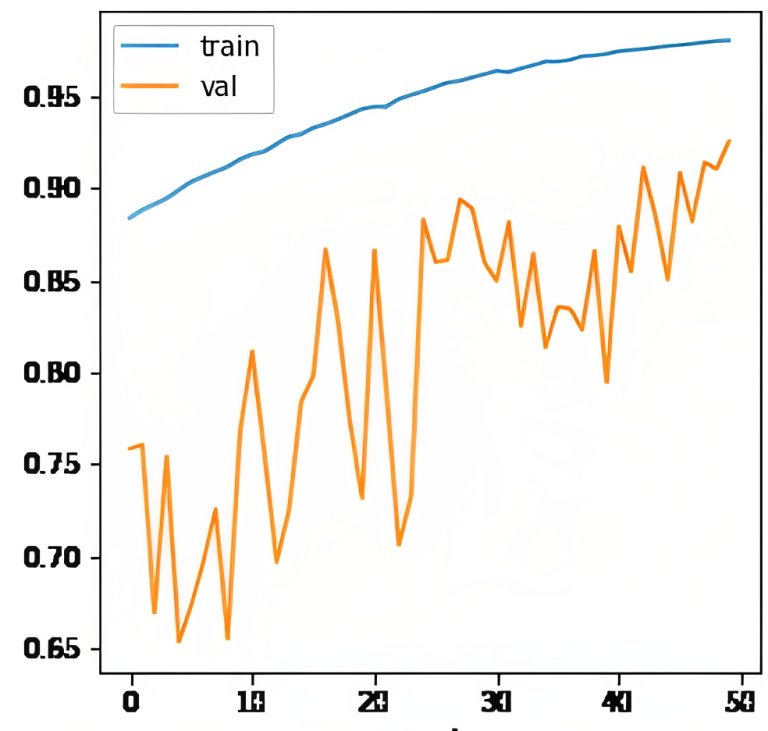}
    \caption{Sensitivity}
    \label{fig:sensitivity1}
  \end{subfigure}
  \begin{subfigure}{0.3\textwidth}
    \includegraphics[width=\linewidth]{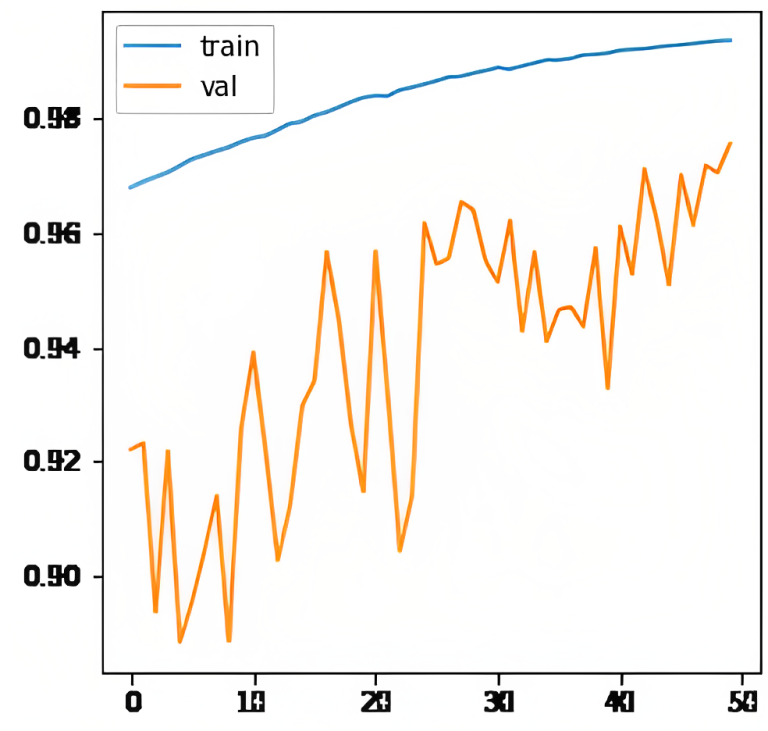}
    \caption{Specificity}
    \label{fig:specificity1}
  \end{subfigure}
  \caption{Performance metrics vs. epochs for the first round }
  \label{fig:performance_metrics1}
\end{figure}

During the initial round of training with a batch size of 16, the attention U-Net model demonstrated steady and promising improvement in performance, as depicted in Figure \ref{fig:performance_metrics1}. The accuracy of the model started at 0.88 and consistently increased, reaching an impressive value of 0.96 at the end of the 50 epochs. Similarly, other evaluation metrics showed continuous progress as well. The dice coefficient and IoU, which assess the agreement between the model's predictions and the ground truth segmentations, also exhibited a positive trend, achieving values of 0.9 and 0.875, respectively, by the end of this round. Additionally, sensitivity and specificity metrics, which measure the model's ability to correctly identify positive and negative cases, also displayed favorable trends throughout the training process. Sensitivity increased from 0.75 to 0.9, while specificity improved from 0.92 to 0.97.
\begin{figure}
  \centering
  \begin{subfigure}{0.3\textwidth}
    \includegraphics[width=\linewidth]{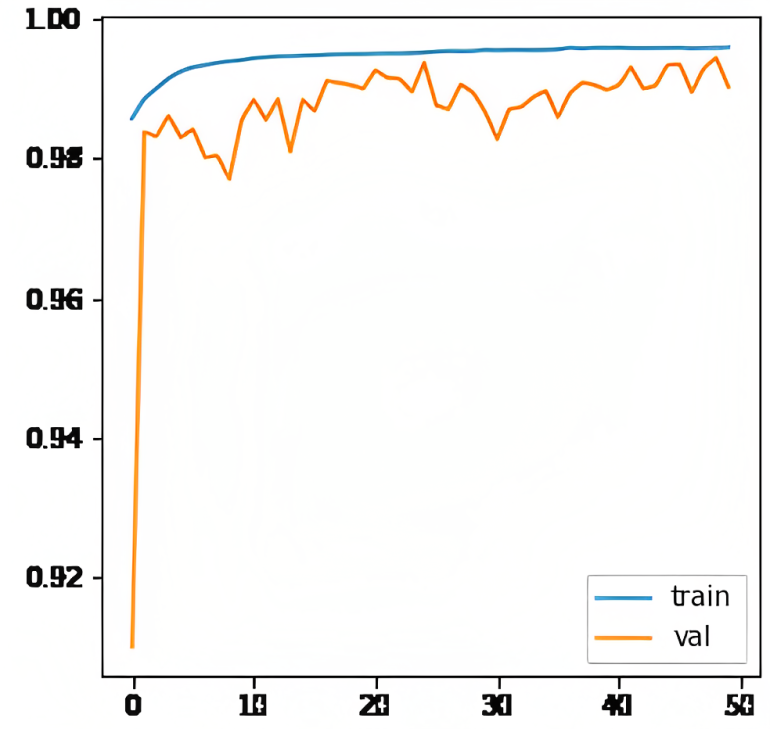}
    \caption{Accuracy}
    \label{fig:accuracy2}
  \end{subfigure}
  \begin{subfigure}{0.3\textwidth}
    \includegraphics[width=\linewidth]{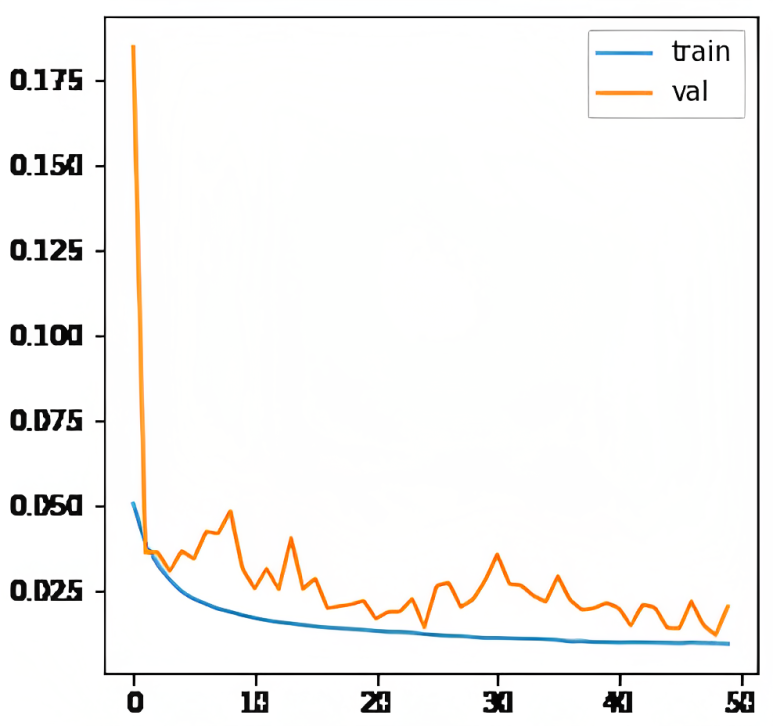}
    \caption{Dice Loss}
    \label{fig:dice_loss2}
  \end{subfigure}
  \begin{subfigure}{0.3\textwidth}
    \includegraphics[width=\linewidth]{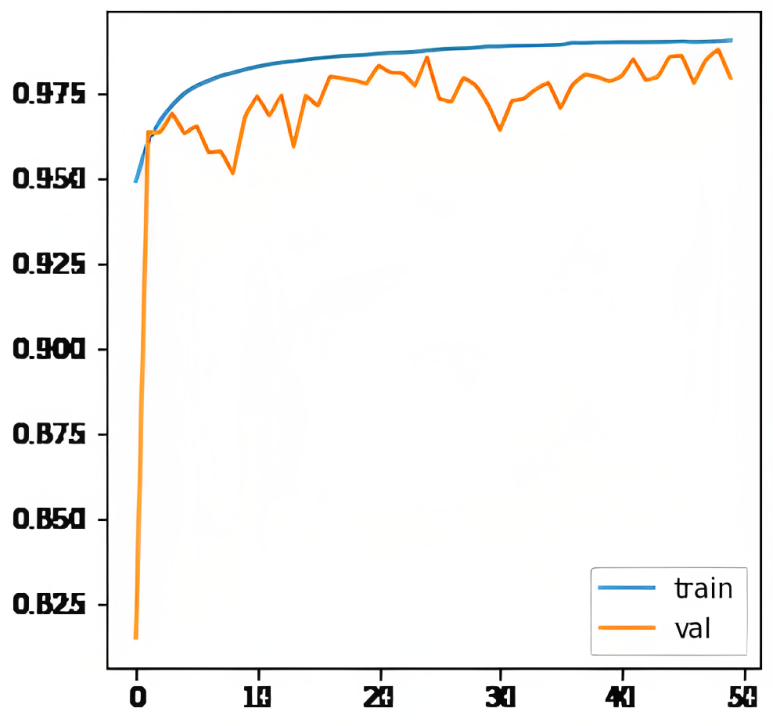}
    \caption{Dice Coefficient}
    \label{fig:dice_coefficient2}
  \end{subfigure}
  \medskip
  \begin{subfigure}{0.3\textwidth}
    \includegraphics[width=\linewidth]{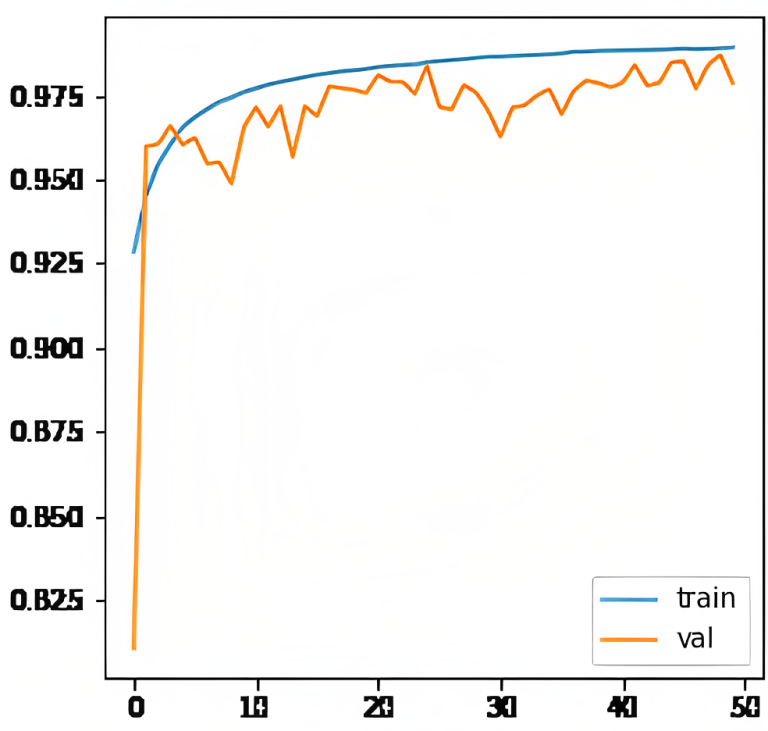}
    \caption{IoU}
    \label{fig:iou2}
  \end{subfigure}
  \begin{subfigure}{0.3\textwidth}
    \includegraphics[width=\linewidth]{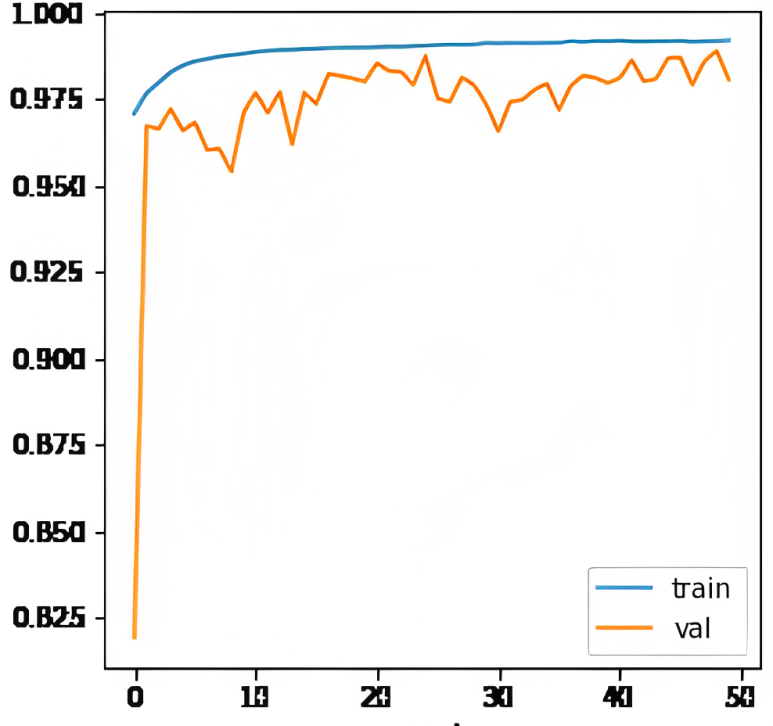}
    \caption{Sensitivity}
    \label{fig:sensitivity2}
  \end{subfigure}
  \begin{subfigure}{0.3\textwidth}
    \includegraphics[width=\linewidth]{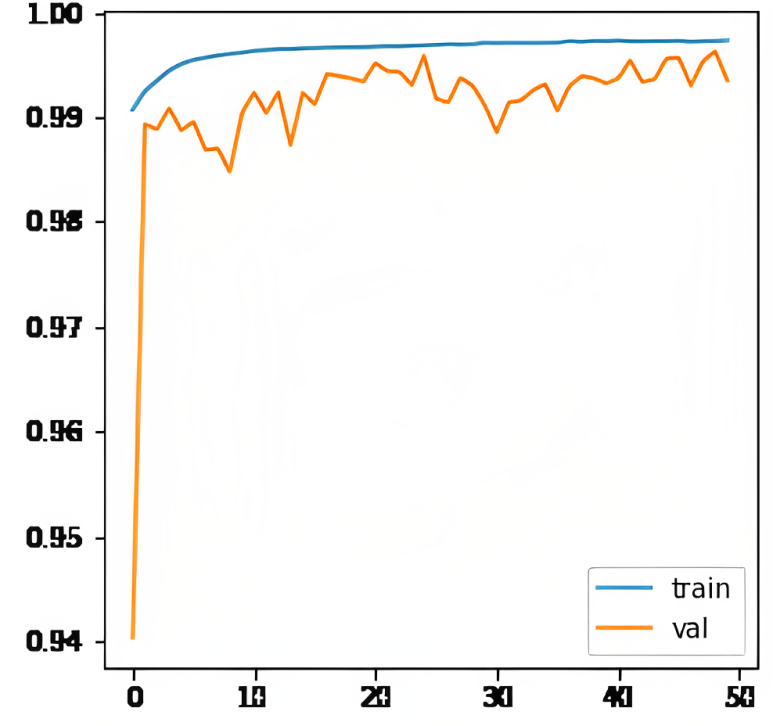}
    \caption{Specificity}
    \label{fig:specificity2}
  \end{subfigure}
  \caption{Performance metrics vs. epochs for the second round}
  \label{fig:performance_metrics2}
\end{figure}

Building upon the progress achieved in the first round, the attention U-Net model's performance further excelled during the second round of training with a batch size of 8. As illustrated in Figure \ref{fig:performance_metrics2}, the model continued to demonstrate exceptional performance across various metrics after an additional 50 epochs. The accuracy remained consistently high at 0.984, showcasing the model's overall proficiency in segmentation tasks. The dice coefficient and IoU both reached nearly 0.98, indicating a significant improvement over the initial scores and suggesting that the model successfully captured tumor regions with higher precision and improved overlap with the ground truth. The loss function also remained consistently low at 0.025, signifying the model's stability and efficiency in minimizing errors. Lastly, sensitivity and specificity scores remained impressively high at 0.975 and 0.996, respectively, further affirming the model's ability to accurately detect tumor and non-tumor regions.

\begin{figure}
\centering
\includegraphics[width=.8\textwidth,height=20pc]{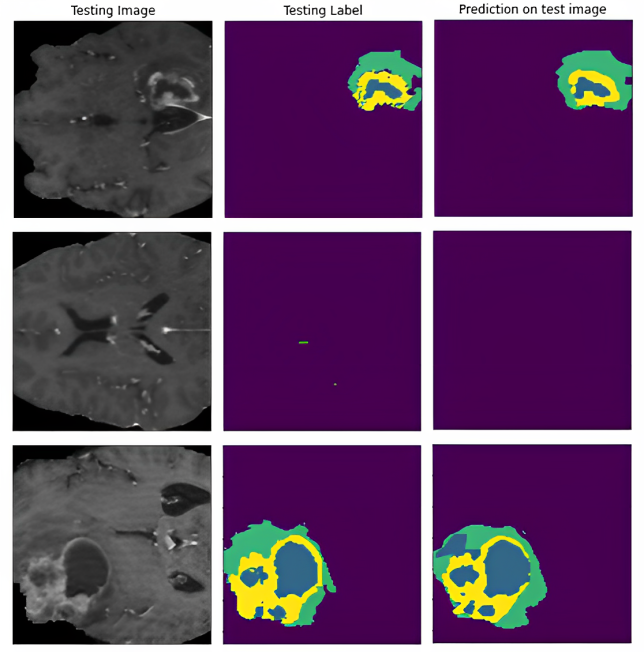}
\caption{Segmentation results on three test scans} \label{fig:viz}
\end{figure}
\section{Discussion}

 By incorporating attention mechanisms, the model effectively focused on relevant features while disregarding irrelevant ones, leading to improved segmentation performance. For a qualitative assessment of the model's performance, a further analysis was conducted by visually comparing the ground truth with the predicted segmentation of three scans, as depicted in Figure \ref{fig:viz}. The visualization of the segmentation results aligns well with the results of the performance metrics discussed earlier, providing evidence of the model's proficiency in accurately capturing true positive regions. Of particular note is the second test (second row), which illustrates the model's effectiveness in correctly identifying true negatives. This aligns with the high specificity observed in the performance metrics, indicating the model's capability to correctly identify negative cases. However, it is worth noting that in the third test (third row), there are some misclassified pixels in the ET region. This can be attributed to the fact that it has the lowest class distribution, making it more challenging for the model to accurately identify and segment these regions.
\begin{table}[H]
\centering
\caption{Comparison of performance metrics with recent studies}
\label{tab:performance_comparison}
\begin{tabular}{|l|c|c|c|c|c|c|}
\hline
\textbf{Study} & \textbf{Accuracy} & \textbf{Sensitivity} & \textbf{Specificity}  & \textbf{Dice Coefficient} \\
\hline
Montaha et al. \cite{Lit1} & 0.994 & 0.989 & 0.997  & 0.939  \\
Ilhan et al. \cite{Lit2} & 0.994 & 0.836 & 0.998 & 0.880  \\
Cinar et al. \cite{Lit3} & N/A & 0.931 & 0.995  & 0.931  \\
Raza et al. \cite{Lit4} & N/A & 0.971 & 0.986 & 0.834  \\
Gab Allah et al. \cite{DisTable1} & N/A & 0.912 & 0.996 &  0.893  \\
Cao et al. \cite{DisTable2} & N/A & 0.870 & 0.996 &  0.852  \\
\textbf{This study} & \textbf{0.992} & \textbf{0.988} & \textbf{0.995} & \textbf{0.975} \\
\hline
\end{tabular}
\end{table}
The proposed model was compared to recent studies, and the results are summarized in Table \ref{tab:performance_comparison}. While this study achieved slightly lower accuracy, sensitivity, and specificity compared to Montaha et al. and Ilhan et al., it demonstrated a higher Dice coefficient, indicating better overlap between predicted and ground truth segmentations. Cinar et al. achieved high sensitivity and specificity but did not report accuracy. Raza et al., Gab Allah et al., and Cao et al. reported good sensitivity but did not provide accuracy or specificity values. Overall, this study showed competitive performance in brain tumor segmentation, particularly with a notable improvement in the Dice coefficient, which reflects overall segmentation accuracy.

\section{Conclusion}
This study explores the integration of the attention mechanism into the 3D U-Net model and a tumor detection algorithm based on digital image processing techniques for brain tumor segmentation. By incorporating the attention mechanism, the model effectively focuses on relevant regions and important features, capturing intricate details and prioritizing informative areas during segmentation. Additionally, the tumor detection algorithm addresses the challenge of imbalanced training data. The evaluation of the proposed model on the BraTS2020 dataset demonstrates its effectiveness, achieving an accuracy of 0.992 and a dice coefficient of 0.975. These results indicate precise tumor boundary delineation with minimal error and reliable localization of tumor regions. This study contributes to brain tumor segmentation research and highlights the potential of advanced deep-learning techniques, such as the attention mechanism, to enhance accuracy and reliability in medical imaging. Further exploration of these techniques promises significant advancements in medical diagnostics.

%

\bibliography{ref.bib}

\end{document}